\documentclass[a4paper, 10pt, twocolumn, twoside]{article}

\usepackage{ISARC}

\usepackage{lscape}
\usepackage{hologo}
\usepackage{todonotes}

\begin{document}

\linespread{0.5}

\title{Are Open-Vocabulary Models Ready for Detection of MEP Elements on Construction Sites}

\author{Abdalwhab Abdalwhab$^{1}$, Ali Imran$^{1}$, Sina Heydarian$^{2}$, Ivanka Iordanova$^{2}$ and David St-Onge$^1$}

\affiliation{
$^1$Lab INIT Robots, Department of Mechanical Engineering, ETS Montreal, Canada\\
$^2$GRIDD, Department of Construction Engineering, ETS Montreal, Canada\vspace{-0.8em}
}

\email{
\href{mailto:abdalwhab-bakheet-mohamed.abdalwhab.1@ens.etsmtl.ca}{abdalwhab-bakheet-mohamed.abdalwhab.1@ens.etsmtl.ca}
}


\maketitle 
\thispagestyle{fancy} 
\pagestyle{fancy}

\begin{abstract}
The construction industry has long explored robotics and computer vision, yet their deployment on construction sites remains very limited. These technologies have the potential to revolutionize traditional workflows by enhancing accuracy, efficiency, and safety in construction management. Ground robots equipped with advanced vision systems could automate tasks such as monitoring mechanical, electrical, and plumbing (MEP) systems. The present research evaluates the applicability of open-vocabulary vision-language models compared to fine-tuned, lightweight, closed-set object detectors for detecting MEP components using a mobile ground robotic platform. A dataset collected with cameras mounted on a ground robot was manually annotated and analyzed to compare model performance. The results demonstrate that, despite the versatility of vision-language models, fine-tuned lightweight models still largely outperform them in specialized environments and for domain-specific tasks.
\end{abstract}

\begin{keywords}
Robotics; MEP Detection; Open-Vocabulary Models
\end{keywords}
\vspace{-1em}

\section{Introduction}
\label{sec:Introduction}
Recent surveys highlight the growing adoption of mobile robots in the construction industry as a means to address labor shortages and improve safety~\cite{carra2018robotics}. Despite these advancements, significant challenges remain for real-world deployment. As noted by \cite{braga2024robotic}, collaboration between developers and construction firms is essential to ensure these technologies contribute effectively to productivity and safety. Among the most repetitive and time-consuming tasks in construction workflows, site surveys—critical for monitoring progress and supporting decision-making—are still largely performed manually~\cite{wang2024future}. Robust detection technologies could significantly enhance the automation of these processes, but detecting Mechanical, Electrical, and Plumbing (MEP) systems remains a complex challenge.




\begin{figure}[t]
    \centering
    \includegraphics[width=0.5\textwidth]{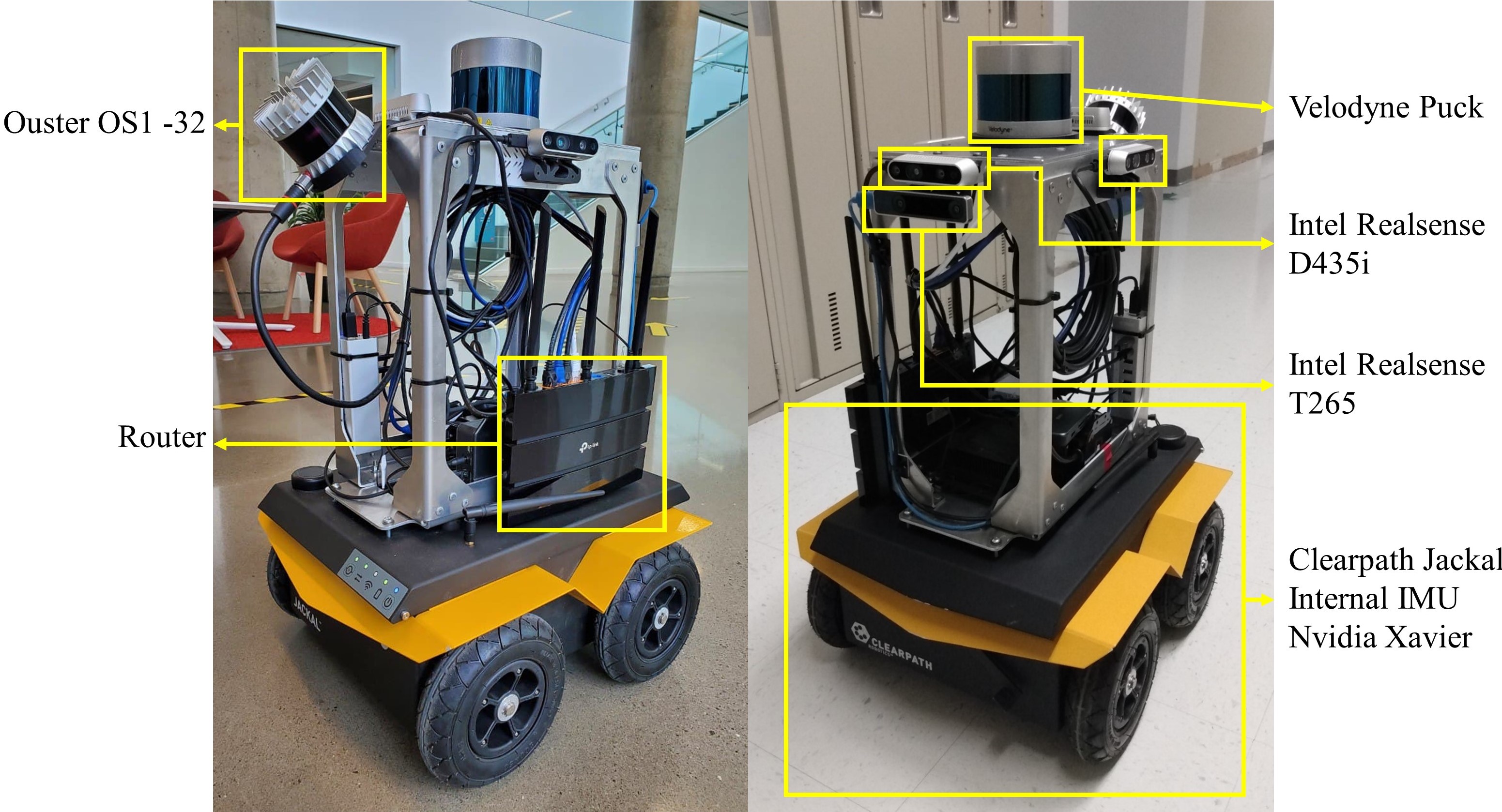}
    \caption{Our Journeybot platform for data collection: a Jackal rover base from Clearpath equipped with several cameras and LiDARs (not used in this study).}
    \label{fig:robotic_platform}
    \vspace{-2em}
\end{figure}

MEP monitoring involves tracking and assessing the installation and functionality of critical systems during construction. This process is essential for identifying potential issues early, preventing costly rework, and avoiding delays~\cite{roh2011object}. While supervised learning algorithms, particularly convolutional neural networks (CNNs), have significantly advanced object detection in construction, these methods face limitations in dynamic and unstructured environments~\cite{park2023small}. Models like YOLO (You Only Look Once) have shown success in controlled conditions but struggle with occlusions, variable lighting, and the introduction of novel objects often encountered on construction sites~\cite{hao2024casnli, diwan2023object}.

Open vocabulary object detection presents an alternative approach, offering the ability to generalize across diverse datasets and identify objects not seen during training~\cite{cheng2024yolo}. These open-vocabulary models, which integrate Large Language Models (LLMs) with computer vision, have demonstrated potential in tasks such as generating detailed inspection reports and monitoring construction sites~\cite{pu2024autorepo}. Another example, Omni-Scan2BIM~\cite{wang2024omni}, employs open-vocabulary models to recognize and segment MEP components for automatic as-built BIM generation. However, their focus on specific classes, such as pipes and ducts, limits their applicability to broader MEP monitoring tasks. Trained on diverse datasets, open-vocabulary models can adapt to new environments and identify emerging components and materials~\cite{salimi2023constscene}, an essential capability for dynamic construction sites where variability is high.

This paper evaluates the potential of open vocabulary object detection models for MEP monitoring in construction environments. To this end, we constructed a real-world dataset of MEP components using a mobile robot equipped with multiple sensors. The dataset was used to train and evaluate a fine-tuned supervised object detection model, YOLO11~\cite{khanam2410yolov11}, which as a closed-set detector works only for the specific classes that it was trained or fine-tuned on. Then, compare its performance with three open vocabulary object detection models: Grounding-SAM2~\cite{ren2024grounded}, Grounding DINO~\cite{liu2025grounding}, and DETIC~\cite{zhou2022detecting}. This comparative analysis provides a comprehensive assessment of fine-tuned, lightweight closed-set detectors versus pre-trained open-vocabulary models in real-world MEP monitoring tasks. Our findings offer valuable insights into the effectiveness of deploying open-vocabulary models compared to smaller, task-specific models for dynamic construction environments. As LLMs and open-vocabulary models continue to evolve, this work contributes to understanding their role in advancing construction site automation.

\section{Experimental Setup}
\label{ExperimentsAndResults} 

\textbf{Robotic Platform}: This study utilized the Journeybot robotic platform, a four-wheeled unmanned ground vehicle (UGV) equipped with a hybrid vision and laser sensing system (Figure~\ref{fig:robotic_platform}). While capable of semi-autonomous navigation, the robot was manually operated during data collection for safety reasons.

The sensing system was tailored for comprehensive environmental monitoring and digital twin data collection. It included four Intel Realsense D435i stereo cameras covering all sides of the robot, with one facing upward, and a front-facing Intel Realsense T265 tracking camera for enhanced localization. Depth images from the D435i supported collision avoidance, together with a horizontally mounted Velodyne Puck 32MR LiDAR. An Ouster LiDAR augmented the data collection system, but both LiDARs are not covered by this work. Wheel encoder data and IMU readings were synchronized with detection sensors by an NVidia Jetson computer.

\textbf{Dataset}: \label{dataset}
The dataset was collected at an active construction site provided by an industrial partner, with a primary focus on MEP assets. Using the Robot Operating System (ROS), our robotic platform was manually teleoperated through the site, capturing data stored as \emph{rosbags}. Images were extracted at a rate of 10 Hz from the rosbags, followed by a filtering process to remove redundant and blurred frames. The remaining images were annotated for the 10 specific MEP components listed in Table~\ref{tab:dataset_Details} using an online annotation tool. The dataset is shared \href{https://universe.roboflow.com/yolo-nas-frb7l/mep-elements-in-construction-site}{online}.



\begin{table}[!htb]
    \vspace{-1em}
    \centering
    \caption{Dataset details with number of images (and instances) per each class.}
    \label{tab:dataset_Details}
    \begin{tabular}{ccc}
    \hline
No. &  Class& Images(instances) \\ \hline
&All&8885(14064) \\
1&Boiler& 768(769)    \\
2&Cable Tray fitting& 252(317)    \\
3&Electrical Panel& 658(1022)    \\
4&Fire Alarm Detector& 872(877)    \\
5&Pipe Fitting&3625(5014)    \\
6&Valve& 610(906)    \\
7&Electrical Outlet&786(1023)    \\
8&Generator& 124(124)   \\
9&Light& 1992(2168)    \\
10&Pump& 1536(1844)    \\
    \hline
    \end{tabular}
\end{table}

The final annotated dataset was divided into 70\% training, 20\% validation, and 10\% testing splits. Table~\ref{tab:dataset_Details} also presents the number of images and instances per class.  Example images are shown in the results section (Figure~\ref{fig:y11_detections}), illustrating the challenges posed by real-world data collection on a mobile platform. These include objects appearing at varying distances, being partially occluded, or blurred due to motion, reflecting the dynamic and unpredictable nature of construction site environments.



\textbf{Evaluation Procedure}: \label{eval_procedure}
For quantitative evaluation, we assessed the performance of a lightweight object detector (YOLO11 Nano pre-trained on COCO~\cite{lin2014microsoft}) fine-tuned (200 epochs) on our dataset, comparing its best performing weights (on the validation split) against three pre-trained large open-vocabulary models without fine-tuning: Grounding DINO, Grounded SAM2, and DETIC. The evaluation metrics included precision, recall, and F1 score (All measured on the testing split at 0.5 intersection over union). 






\section{Results and Discussion}


Table~\ref{tab:F1Score} depicts the results of F1 score for YOLO11 Nano compared to the open-vocabulary models GSAM2, GDINO, and DETIC, evaluated on the test dataset. YOLO11 achieved a total precision of 0.87, outperforming GSAM2, GDINO, and DETIC, which scored 0.018, 0.02, and 0.015, respectively. Similarly, YOLO11 exhibited superior recall at 0.901, compared to 0.042, 0.084, and 0.018 for the open-vocabulary models.

\begin{table}[!htb]
    \vspace{-1em}
    \centering
    \caption{F1 score total and per class for each model evaluated in the testing dataset split}
    \label{tab:F1Score}
    \begin{tabular}{ccccc}
    \hline
Class&YOLO11&GSAM2&GDINO& DETIC \\ \hline
all&0.89&0.018&0.032&0.014 \\
1&0.94&0.01&0&0.005 \\
2&0.86&0&0&0 \\
3&0.9&0.045&0.059&0.043 \\
4&0.94&0.036&0.026&0.063 \\
5&0.86&0.01&0.013&0.013 \\
6&0.87&0.005&0.007&0 \\
7&0.82&0.005&0.003&0 \\
8&0.92&0&0.006&0 \\
9&0.81&0.038&0.027&0 \\
10&0.93&0.014&0.123&0.013 \\
    \hline
    \end{tabular}
\end{table}

The results underscore the superior performance of YOLO11 Nano, with improvements exceeding 85\% in precision, 82\% in recall, and 86\% in F1 score compared to the best-performing open-vocabulary model (GDINO). YOLO11 also achieved consistent detection across all classes, including class 2 (Cable Tray Fitting), which none of the open-vocabulary models detected. The failure of these models in detecting such specialized classes likely stems from their reliance on large, generic datasets like COCO~\cite{lin2014microsoft} and LVIS~\cite{gupta2019lvis}, which lack adequate representation of MEP-specific components. DETIC exhibited the poorest performance, failing in five out of ten classes, while GSAM2 and GDINO each failed in two classes.

\begin{figure}[!htb]
    \centering
    \includegraphics[width=\columnwidth]{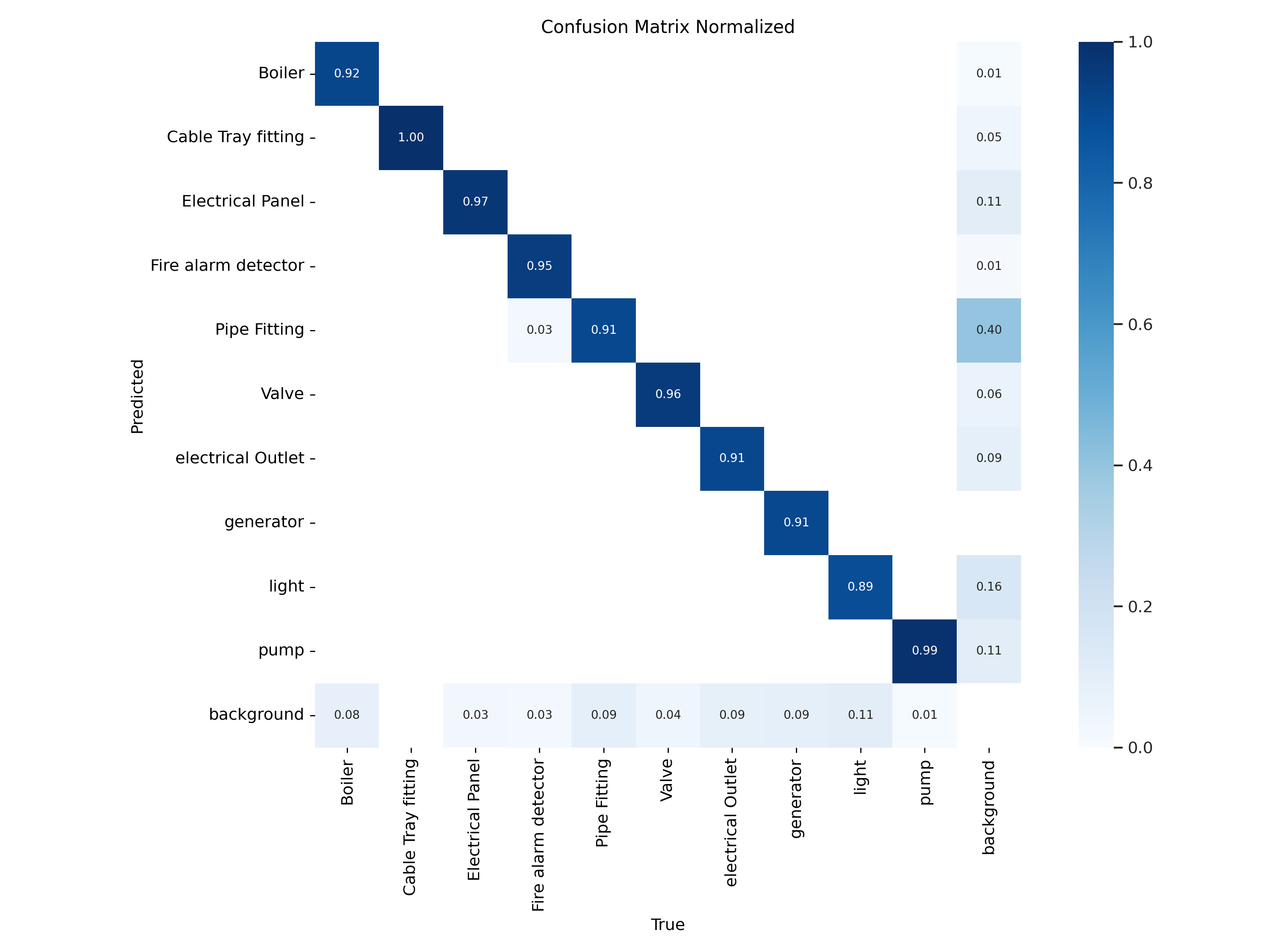}
    \caption{YOLO11 normalized confusion matrix for its predictions on the testing dataset split.}
    \label{fig:y11_confusion}
    \vspace{-1em}
\end{figure}

Figure~\ref{fig:y11_confusion} presents the normalized confusion matrix for YOLO11 Nano's predictions, demonstrating robust performance even for less-represented classes, such as generators, with a detection accuracy of 91\%. This emphasizes YOLO11's capability to handle imbalanced datasets, a common challenge in domain-specific applications.

\begin{figure}[!htb]
    \centering
\includegraphics[width=0.48\textwidth]{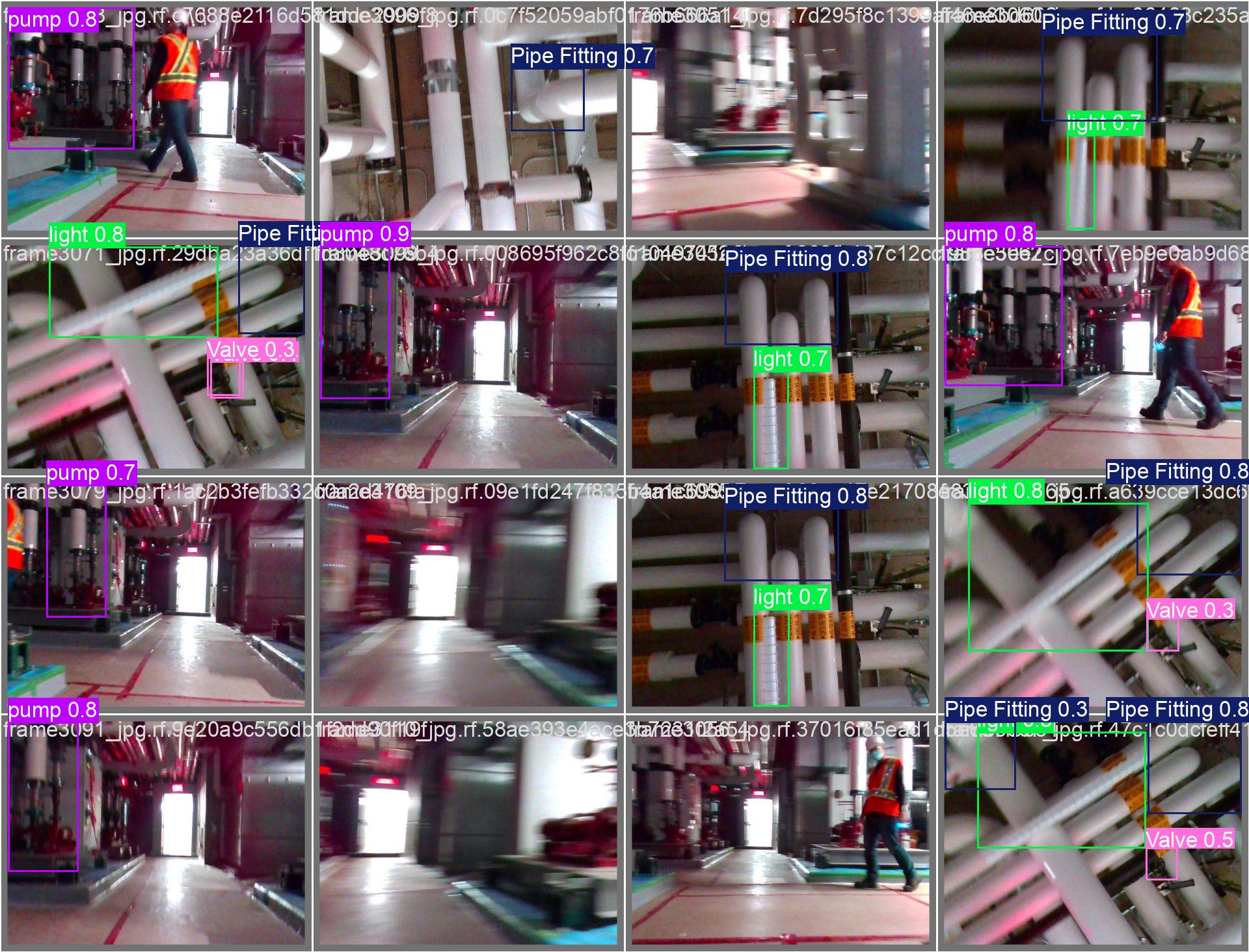}
    \caption{Qualitative examples of YOLO11 Nano's performance, displaying predicted classes, bounding boxes, and confidence scores. }
    \label{fig:y11_detections}
    \vspace{-2em}
\end{figure}

Figure~\ref{fig:y11_detections} provides qualitative examples of YOLO11 Nano's predictions, displaying predicted classes, bounding boxes, and confidence scores. In some cases, the model outperformed human annotations, identifying overlooked objects like valves in the bottom-right corner of an image. Conversely, in instances of blurred images, the model avoided false positives, aligning with human annotators' limitations. However, false detections occasionally occurred, particularly for reflective surfaces, which resembled target classes like light sources.
 
These findings highlight the limitations of general-purpose open-vocabulary models in specialized environments like construction sites, where dynamic conditions, specialized components, and unique visual contexts demand fine-tuned, domain-specific approaches.

\textbf{Real-world Applicability}: Assessing the applicability of open-vocabulary models for real-world robotic systems compared to lightweight object detectors like YOLO11 Nano requires careful consideration of model size and memory requirements. Model size, expressed in terms of parameters, is a critical factor influencing memory usage and computational efficiency, especially on resource-constrained platforms.
Among the evaluated models, GSAM2 exhibited the largest memory footprint with 910 million parameters—350 times larger than YOLO11 Nano, which has just 2.6 million. GDINO followed, with a parameter count 67 times larger than YOLO11 Nano, while DETIC, the smallest open-vocabulary model, required 164 million parameters, making it 63 times larger than YOLO11 Nano.


Computational efficiency is equally crucial for deployment on real-time robotic systems. YOLO11 Nano was specifically evaluated on an NVIDIA Jetson Orin Nano, processing image samples from the testing dataset at a speed of 23.36 frames per second. This performance meets the requirements for real-time applications, demonstrating that YOLO11 Nano is highly suited for resource-constrained environments, unlike larger open-vocabulary models whose significant computational demands make them less practical for such applications.

\section{Conclusion}
This study compared the performance of pre-trained open-vocabulary detection models and fine-tuned lightweight detectors for robotic MEP detection in construction sites. Using a dataset collected with a tele-operated robotic platform, we demonstrated that fine-tuned models like YOLO11 Nano significantly outperformed open-vocabulary models in detection accuracy, computational efficiency, and real-time applicability. While open-vocabulary models offer promising generalization capabilities, their performance on domain-specific tasks remains limited.

With the huge advancements in vision-language models, we hope this work encourages research on narrowing this gap, possibly, by fine-tuning open-vocabulary models on specialized datasets, potentially combining their adaptability with the precision of supervised models. Another direction could be to design construction-related prompts to adapt the open-vocabulary detectors.


\bibliography{ISARC}

\end{document}